# BERT: A Review of Applications in Natural Language Processing and Understanding

Koroteev M.V., Financial University under the government of the Russian Federation, Moscow, Russia
mvkoroteev@fa.ru

*Abstract*: In this review, we describe the application of one of the most popular deep learning-based language models - BERT. The paper describes the mechanism of operation of this model, the main areas of its application to the tasks of text analytics, comparisons with similar models in each task, as well as a description of some proprietary models. In preparing this review, the data of several dozen original scientific articles published over the past few years, which attracted the most attention in the scientific community, were systematized. This survey will be useful to all students and researchers who want to get acquainted with the latest advances in the field of natural language text analysis.

*Keywords*: natural language processing, review, BERT, language models, machine learning, deep learning, transfer learning, NPL applications,

## Introduction

The search for a universal representation of text is at the heart of the automated processing of natural languages. The big breakthrough in this area has been with the development of pretrained text attachments such as word2vec [52] or GloVe [64]. Over the past years, supervised models have shown consistently better results than unsupervised models [49]. However, in recent years, models based on learning without a teacher have become much more widespread since they do not require the preparation of a specially labeled dataset, but can use already existing or automatically generated huge corpora of texts and, as a result, learn on much a larger sample, thus taking full advantage of deep learning.

The centerpiece of 2019 in the field of natural language processing was the introduction of a new pretrained BERT text attachment model, which enables unprecedented precision results in many automated word processing tasks. This model is likely to replace the widely known word2vec model in prevalence, becoming, in fact, the industry standard. Throughout 2019, almost all scientific articles devoted to the problem of word processing in natural languages, in one way or another, were a reaction to the release of this new model, the authors of which have become one of the most cited researchers in the field of machine learning.

Natural language processing tasks include a wide range of applications from conversational bots and machine translation to voice assistants and online speech translation. Over the past few years, this industry has experienced rapid growth, both quantitatively, in the volume of market applications and products, and qualitatively, in the effectiveness of the latest models and the proximity to the human level of language understanding.

One of the central themes in natural language processing is the task of text representation. Text representation is a kind of rule for converting natural language input information into machine-readable data. A representation can also be considered simply a computer encoding of text, but in the context of applied machine learning problems, such representations that reflect the internal content and conceptual structure of the text are more useful.

The most simple textual representations are categorical encoding when each word is represented as a vector filled with zeros everywhere, except for one position corresponding to the number of this word in the dictionary. This concept was used in the early stages of the industry. It is quite simple, does not require computational resources to implement, and conceptually very simple. However, such a representation does not take into account the semantic features of words, it is rather voluminous and redundant since a vector with the dimension of the number of words in the dictionary is used to represent each word.

A similar view is the well-known bag of words model. This model represents the entire text as a vector with the dimension of a vocabulary, in which each component represents the number of occurrences of a given word in the text. This is also a fairly simple model that does not take into account the semantics of words, however, it is quite successfully applied to tasks, for example, the categorization of texts.

The most applicable in the modern industry is the representation of text in the form of so-called attachments - a mechanism for representing each word in the form of a vector, each coordinate of which has a certain semantic meaning. Most often, attachments with hundreds of coordinates are used. Attachments can capture the semantic meaning of specific words in a text and display it as a coordinate in multidimensional space. One well-known illustration of this is the ability to perform vector operations in this semantic latent space.

Nesting is most often done by unsupervised model training on large text corpora. An example of such tasks can be filling in gaps in the text, determining the relevance of sentences. Learning textual representations is a very computationally intensive task. For most problems of analysis of general-purpose texts, ready-made representations, trained in advance, are used [70].

**Basic concepts about the new BERT linguistic model**

Today, the most advanced text models use transformers to teach how to represent text. Transformers are a type of neural network that are increasingly finding their use in various branches of machine learning, most often in sequence transduction problems, that is, such problems when both the input and output information is a sequence [22]. Such models use a combination of recurrent and convolutional neural networks. Regular recurrent networks perform rather poorly with long-range context. However, in natural text, the representation of the token can be influenced by the context through several words and even sentences from the token itself. To take into account the long-range influence, LSTMs are used in conjunction with the attention mechanism to improve the learning efficiency, taking into account the influence of distant tokens [19].

At the end of 2018, a group of scientists from the Google AI Language laboratory under the leadership of J. Devlin presented a new linguistic model called BERT [16]. This model is intended for deep preliminary learning of bidirectional text representation for subsequent use in machine learning models. The advantage of this model is its ease of use, which involves adding just one output layer to the existing neural architecture to obtain text models that surpass the inaccuracy of all existing ones in several natural text processing problems.

There are two categories of natural text processing tasks: holistic, operating with text at the sentence level, and tokenized ones, such as answering a question and attribution of entities, which produce more detailed output at the level of individual text elements. Both categories of problems have recently been using pretrained models, which can significantly reduce the time for designing and training private models while maintaining a high level of efficiency[14,24].

Regarding the approaches to pre-training models of deep text representation, two approaches are also distinguished: feature extraction and fine-tuning of models. Both approaches use the same pre-learning objective functions and unidirectional text analysis. The first, implemented, for example, in the ELMo model [65], is to use models with a task-specific architecture to train a certain representation, which will then be used as additional features in applied models. The second involves the construction of models that do not use parameters specific to a particular task but training the model, which is then supposed to be retrained by adjusting all the internal parameters of the model. This approach is used, for example, in the OpenAI GPT model [66].

The authors of the BERT model point out that a significant limitation of existing approaches is their focus, which narrows the choice of possible neural network architectures. For example, the GPT model uses left-to-right text scans so that the representation of each element of text (token) can only take into account the previous, but not subsequent tokens. This may not be optimal for holistic text analysis, but for tokenized analysis, this approach significantly reduces the semantic power of the model, since the meaning of a word can depend on its context, both left and right.

BERT tries to get around this limitation by using learning according to the so-called "masked language models", that is, the target function of learning a given representation formalizes the task of predicting a randomly selected and masked word in a text, taking into account only the surrounding context. Thus, a deep bi-directional transformer is trained.

The BERT model training process includes two stages: pre-training on unlabeled data, and additional training on labeled data for a specific application problem. Depending on the task, the retraining process and the architectures used may differ, although they are all based on the same model with the same set of parameters.

The BERT architecture is based on the multilayer bidirectional transformer described in 2017 by A. Washwani in [80]. The authors trained two versions of the neural network - a standard one with 12 layers and 768 coordinates in the view (110 million trained parameters in total) and a large one with 24 layers and 1024 coordinates (340 million parameters). BERT uses text embeddings described in 2016 to represent an input sequence [85]. A sequence is an arbitrary set of contiguous text tokens. For example, the model uses the same representations for one sentence and for a pair, which allows BERT to be used for a wide range of tasks.

Using the described process of additional training for a specific task, BERT was tested on several standard datasets to compare its performance with other published models. So, on the GLUE test (General Language Understanding Evaluation [83], a set of tasks and datasets that test natural language comprehension), the BERT-based model showed an average superiority of 4.5% and 7% (for standard and large neural networks, respectively) compared to the best-known models. The father-in-law of SQuAD (The Stanford Question Answering Dataset [67], the model is asked a question and a passage of text, the model must choose from the text the passage that answers the question) also shows the superiority of BERT over all existing models (according to the F1 metric, the best model showed 78.0; BERT - 83.1; human result - 89.5). And the SWAG test (The Situations With

Adversarial Generations[91], a set of questions with four possible answers to choose from) shows an accuracy of 86.3, which is higher than that of a human expert (85.0).

In the field of teaching linguistic models without a teacher, two main areas can be distinguished: autoregressive models and autoencoders. Both those and others have their significant disadvantages. For example, autoregressive models are inherently unidirectional and cannot take into account the general environment of a particular token. Autoencoder-based models such as BERT do not have this drawback but rely on artificial text tokens such as [SEP] or [MASK] in their learning process, which creates a contradiction between pre-training and additional training for a specific task, where such tokens are absent ...

A group of scientists led by J. Young from Carnegie Melon University [87] tried to combine these two approaches, proposing the XLNet model. Moreover, the obtained model gives better results in comparison with BERT on a wide range of linguistic problems, such as GLUE [83], SQuAD [67], and RACE.

## BERT Retraining Methodology for TextProblems Text

Classificationclassification is a classic natural language processing problem. It consists of assigning one of a predefined set of categories to a given text sequence. Currently, machine learning systems for text classification use convolutional neural networks [29], recurrent models [42], and attention mechanisms [88]. Also in this area models like word2vecvery effective [52], pretrained, GloVe [64], and ELMo [65] are. Such models avoid the computationally complex initial training of the model on a large body of texts.

For text classification tasks, the BERT model is also used, which allows additional training of the basic model for a specific task using just one additional layer of neurons. The use of BERT makes it possible to obtain models with the currently best performance in text classification problems after additional training according to a certain methodology [74].

There is a growing interest in approaches that make it possible to use knowledge from related problems in word processing problems. Two groups of such methodologies can be distinguished: the use of pre-trained models (transfer learning) and multitasking training. BERT, like earlier models, belongs to the first category. Multitasking learning [8] is also a promising, although not the newest, line of research. Certain results were obtained [41,69] in the joint training of textual representations and the target problem within the same model.

The basic BERT model contains an encoder with 12 transformer blocks, 12 attention areas, and a textual representation dimension of 768. The model receives a text sequence of no more than 512 tokens as input and outputs its vector representation. The sequence can contain one or two segments, separated by a special token [SEP], each of which necessarily begins with the token [CLS], with a special classification representation.

When adapting BERT to specific word processing tasks, a special retraining technique is required. Such techniques can be of three types:

**Further pre-**training: BERT is initially trained on general-purpose texts, which may have a different data distribution than the texts of the target application area. Naturally, the desire to retrain BERT on a specific text corpus can be based on an intra-task dataset (the same dataset that will be used to train the target model), an intra-subject dataset (a set of text data obtained from the same subject area) or a cross-subject dataset, in depending on the nature of the texts and the availability of data sets.

Studies of the effectiveness of BERT retraining [74] show that, first, retrained models show significantly better results compared to models without retraining, and, secondly, intra-subject learning is generally more effective than intra-task learning. Cross-subject learning does not significantly improve performance relative to the original BERT model, which is logical given that BERT is trained on a general set of texts.

**Retraining Strategies**: There are many ways to use BERT for a target. For example, you can use the representation provided by BERT as additional or basic characteristics in the classification model, you can use the inner layers of BERT to get information about the text. Different layers of the neural network can display different levels of syntactic and semantic information of the text [24]. Intuitively, the earlier layers contain more general information. Accordingly, the problem arises of choosing the required layer for inclusion in a specific model.

During additional training of models, the problem of so-called "catastrophic forgetting" often arises [51], which is that in the process of additional training on a specific set of data, knowledge expressed in the form of model weights trained at the stage of preliminary training is quickly erased. This leads to the leveling of the use of pre-trained models. It has been shown that using low learning rates can overcome this problem. The study described above used a learning rate of about 1e-5. Higher speed (4e-4) leads to the fact that training on the target dataset does not converge.

**Multitasking Learning**: In the absence of pre-trained natural language models, multitasking learning is effective in leveraging shared knowledge of multiple target tasks. When researchers are faced with several word processing tasks in the same subject area, it makes sense to retrain the BERT model in these tasks simultaneously.

**The problem of improving the subject-specific classification of texts using BERT**

Traditional text embedding models represent each word of the input text (token) as a numerical multidimensional semantic vector (embedding) following the assumption that words with similar semantic meaning usually appear in similar contexts. Accordingly, the embedding of each word is calculated based on its environment within a certain window size. The result is a kind of context-independent representation of each text token. This approach has certain characteristics that represent the challenges facing word processing experts and machine learning architects [90].

A challenge to ambiguity: In traditional text attachments, each token is represented as a fixed vector representation, regardless of the context in which the word appears in the text. However, in natural languages, a large number of words are polysemantic, that is, they carry completely different semantic loads depending on the context. In the problems of text classification, the differentiation of such semantic differences may be necessary, especially with indicative words.

Subject-specific challenge: when using most traditional textual models, the classification efficiency strongly depends on the architectures of neural networks, which are usually built specifically for each specific subject area, which gives rise to the need for the manual search for the model architecture and the intensive process of model training.

To overcome the first challenge, so-called contextual text models have been developed, for example, CoVe [50] or ELMo [65], which produce different vector representations, depending on the environment of a particular token.

The CoVe model uses a deep encoder with LSTM cells in conjunction with an attention mechanism trained on machine translation tasks. The practice of using this model shows that adding context does significantly improve the efficiency of text models in many word processing tasks.

The ELMo model is based on the representation of text from a bi-directional neural network with memory trained on the tasks of searching for a language model on a huge corpus of texts. Both approaches successfully generalize traditional textual representations to context-sensitive models and are used in specific tasks as sources of additional features that are input to specific models.

Over the past few years, the efforts of researchers have been aimed at creating a general (universal) text model, pre-trained on a large corpus of general-purpose texts, suggesting limited additional training for solving specific word problems. This approach is designed to solve the problem of finding specific model architectures for each problem and, as a result, to significantly reduce the number of trained model parameters. Such models include ULM-FiT [24], OpenAI GPT [66], and BERT.

It is the efficiency of using BERT today that is significantly higher than other text attachments. However, the researchers note that the full potential of BERT as a universal text model has not yet been revealed and point to the main directions of promising research:

- development and detailed analysis of techniques for retraining BERT for specific word problems. For example, the authors in [74] consider techniques for retraining a model on long text sequences consisting of more than 512 tokens.
- a methodology for retraining BERT on small sets of texts, which are not enough to highlight full knowledge of the textual features of the subject area. Since manual data markup is still a very time-consuming process, in many specific areas there is not enough marked-up text selection to fully retrain the text model.
- developing a methodology for unsupervised learning on a large unlabeled corpus of subject-specific texts can be another way to increase the flexibility and universality of BERT for different subject areas.

**Using BERT for Text Annotation Tasks**

Another interesting problem in natural language processing is text annotation - the automatic selection of key phrases and sentences that best characterize the content of a given text. Text annotation methods are widely used in education. An example is a project to manually annotate publicly available transcriptions of MIT lectures [20,51]. As the researchers note, for a limited set of data, such an approach has the right to exist, however, given the ever-increasing volume of information, it is not possible to continue to rely on hand tools. There are two approaches to annotating text - abstractive and extractive. The first imitates the human approach to natural language processing - the use of a wide vocabulary, highlighting key ideas in a compressed volume. This approach seems to be more desirable and is the subject of a large amount of research [18], but it requires enormous computational power and time to process deep machine learning models or to compile complex algorithms and rule bases.

In contrast, the extractive approach uses syntactic structures, sentences, and phrases contained in the text, manipulating only the input content. The use of the manual annotation method throughout the 2000s is associated with the poor quality of the extractive techniques that existed at that time. For example, the model created in 2005 by G. Murray [55] for annotating the minutes of corporate meetings using probabilistic models gave incomparably worse results than human text processing.

Several attempts have been made to build a better algorithm using rhetorical information [92], TextRank ranking algorithms [79], a naive Bayesian classifier [4], or TF-IDF metrics [97]. These methods have been used with varying success, but what they all have in common is that they don't all use deep learning. There is a methodological reason for this: over the years, the most effective method of word processing using machine learning has been recurrent neural networks, which entails the need to use huge data sets for training over many machine hours to achieve an acceptable result. And even so, such models gave low efficiency on long text sequences and were prone to overfitting.

The situation also changed with the introduction in 20017 of special neural network architecture - a transformer [80], which shifted the focus from recurrent and convolutional networks to feedforward networks with an attention mechanism. That is why the widespread introduction of transformers into the practice of text processing in natural languages, the presentation of the pretrained BERT model gave such an impetus to the development of the field of word processing in all tasks.

For text annotation tasks, the use of text attachments gives another non-obvious advantage - the ability to create variable-length annotations. Since the nesting produces a vector representation of the text, this representation can be clustered with an arbitrary size K, which allows building a dynamic presentation of the text [53].

Evaluating the performance of text annotation models is complicated by the fact that there is no generally accepted standard for annotation, which means that there are no automated metrics for assessing the quality of annotation. The only way is to manually assess the quality of the resulting text.

The use of BERT improves the perceived quality of text annotation, although such models are not free from classic drawbacks, such as the difficulty in separating indicative sentences that are a small proportion of the original text. Thus, annotating long texts is still a promising area of research, along with the development of model datasets and quantitative metrics for evaluating the effectiveness of annotation models.

## BERTScore: BERT Based

Text Generation Quality Assessment The problem of text generation quality assessment arises in the process of solving many problems, for example, machine translation. It can be reduced to comparing a set of candidate proposals with a sample proposal in a given textual context. However, the most commonly used sentence similarity metrics, such as the one described in the 2002 BLEU (bilingual evaluation understudy) [63,72], focus only on superficial similarity. The aforementioned BLEU metric, the most common in the development of machine translation systems, relies on the comparison of the intersection of n-grams of text. For all its simplicity, such metrics miss the lexical and semantic diversity of natural languages.

A vivid example of this problem, described in 2005 by B. Sataneev in [5]: several popular text metrics for this reference sentence "People love foreign cars" prefers the sentence "People like to travel abroad" over those that are more semantically close to the original " Customers prefer foreign cars ". As a consequence, machine translation systems that use such metrics to assess translation quality will prefer syntactically and lexically similar constructs, which is suboptimal in the context of wide linguistic diversity.

The representation of the BERT system allows it to be used as a basis for measuring the similarity of sentences in natural languages, using the metric of the distance between text attachments of compared sentences.

In general, the text metric, or the metric of the quality of text generation, is a function $f(x, \hat{x}) \in \mathbb{R}$, where $x \rightarrow <x_1, x_2, ..., x_k>$ is the vectorized representation of the sample proposal, and $\hat{x} \rightarrow <\hat{x}_1, \hat{x}_2, ..., \hat{x}_l>$ is the vectorized representation of the candidate proposal. A good metric should reflect the person's judgment as closely as possible, that is, show a high correlation with the person's assessment results. All existing metrics can be roughly classified into one of four categories: n-gram match, edit distance, attachment comparison, and trained functions.

A good metric should reflect the person's judgment as closely as possible, that is, show a high correlation with the person's assessment results. All existing metrics can be roughly classified into one of four categories: n-gram match, edit distance, attachment comparison, and trained functions.

The most common textual metrics are based on counting the number of n-grams found in both sentences. The larger the dimension n in an n-gram, the more the metric can capture the similarity of whole words and their order, but at the same time, the more this metric is limited and tied to a specific formulation and word forms. The already mentioned BLUE and MeTEOR belong to this category of metrics.

Some methods calculate the proximity of proposals by the number of edits that translate a candidate into a reference. Methods such as TER [72] and ITER [61] take into account the semantic proximity of words and the normalization of grammatical forms. This can also include methods like PER [77] and CDER [37], taking into account the permutation of text blocks. Some more modern methods (CharacTER [84], EED [73]) have unexpectedly better results.

In recent years, metrics have begun to appear based on the use of dictionary embeddings, that is, trained dense word vectorizations (MEANT [44], YISI-1 [46]). The advantage of using BERT as a basis for constructing such metrics is that it takes into account the context of a word within its environment, which makes it possible to use attachments not at the level of individual words, but the level of the entire sentence.

Since the criterion for the quality of a text metric is its correlation with human judgments, it is not surprising that machine learning is used to build trained metrics, where the target function is just such a coincidence. For example, BLEND [48] uses a regression model to weigh 29 known metrics. The disadvantage of this approach is the dependence on the presence of a corpus of pre-labeled data for the training set, as well as the risk of overfitting in a certain subject area and, as a consequence, the loss of generalizing ability and universality.

In February 2020, researchers from Cornwell University proposed a special mechanism for evaluating the effectiveness of text models based on BERT - BERTScore [93]. BERTScore is used to automatically assess the quality of natural language text generation. The authors of the proposal argue that BERTScore correlates better with human judgments about the quality of the text and, as a result, can form the basis for a more efficient and effective model selection process. Today it is the most progressive metric for evaluating the quality of text generation.

At the heart of BERTScore is a simple algorithm for calculating the cosine distance between vectorized representations of each word in a reference sentence and a candidate sentence. Distances are calculated in pairs, and then, for each word in the reference, the closest words in the candidate are selected, these distances are averaged

$$R_{BERT} = \frac{1}{|x|} \sum_{x_i \in x} \max_{\hat{x}_j \in \hat{x}} x_i^T \hat{x}_j$$

and make up an estimate - a review score. Accuracy score is considered

$$P_{BERT} = \frac{1}{|x|} \sum_{\hat{x}_j \in \hat{x}} \max_{x_i \in x} x_i^T \hat{x}_j$$

similar, but inverted relative to candidate and reference - . In addition, an

F1 score is calculated -
$$F_{BERT} = 2 \frac{P_{BERT} \cdot R_{BERT}}{P_{BERT} + R_{BERT}} \quad R_{BERT} = \frac{1}{|x|} \sum_{x_j \in x} \max_{\hat{x}_j \in \hat{x}} x_i^T \hat{x}_j$$
. - review
score. Accuracy score is considered similar, but inverted relative to candidate and reference -

$$P_{BERT} = \frac{1}{|x|} \sum_{\hat{x}_j \in \hat{x}} \max_{x_i \in x} x_i^T \hat{x}_j$$
. Also, an F1 score is calculated -
$$F_{BERT} = 2 \frac{P_{BERT} \cdot R_{BERT}}{P_{BERT} + R_{BERT}}$$
.

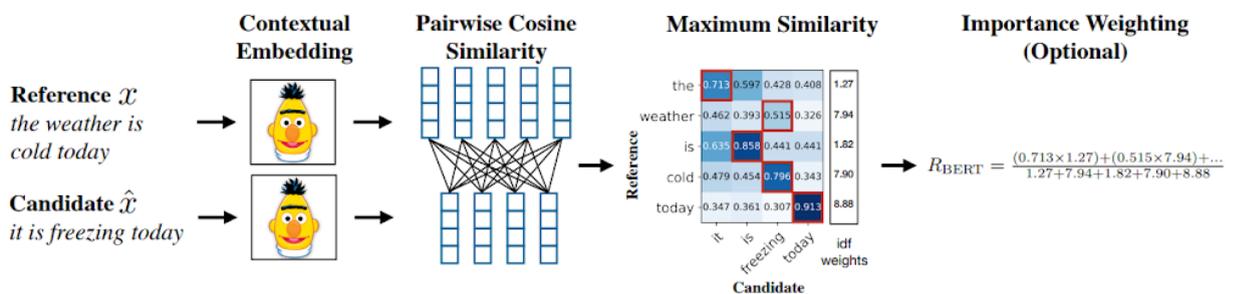

Source: [93]

In addition to the basic algorithm, the authors of BERTScore use the IDF metric to determine the rarer words, as previous studies of text metrics [81] indicate that sparse words may be more indicative of the similarity of two sentences. The IDF metric estimated on the reference and smoothed (+1) to account for new words is used as the weight of the corresponding cosine measure when averaged.

In a comparative analysis of text similarity metrics, metrics based on BERT show consistently higher results than classic text metrics. This means that they are statistically significantly closer to human estimates.

Besides, the original article introducing BERTScore focuses on performance issues. In this regard, BERTScore is, of course, often slower than classic models. The authors give an assessment of the comparison with the popular implementation of SacreBLEU, in which BERTScore is about three times slower. Since the pretrained BERT model is used to assess the similarity, the increase in the accuracy of the estimate is given at the expense of a decrease in speed that would not be expected from such a more complex model. Given the typical size of natural text processing datasets, the increase in computation time during model validation should not have a significant effect on the performance of the machine learning system as a whole.

Publications already beginning to appear [95] that improve the original BERTScore algorithm by parallelizing computational operations. Variations of classical metrics [45]using BERT and showing improved correlation with human estimates are presented.

Undoubtedly, the direction of the future development of textual metrics will be the wider use of BERT as a basis for assessment, a kind of semantic engine. Also promising is the development of specific models that take into account the peculiarities of specific subject areas and increase the basic level of accuracy through specialization. Concerning BERTScore, another advantage is its differentiability, which will allow it to be integrated into the methodology for training text models in the future, which promises to further increase the performance and quality of machine learning models of natural text processing.

**BERT Based Attacks on Text Classification**

Models Machine learning models can very often be vulnerable to input data that are indistinguishable from a real example for humans, but still, give an incorrect output [32]. Such specially selected (adversarial) examples are correctly classified by a human expert, but give an incorrect result in the model, which casts doubt on the safety and reliability of the applied machine learning models [62]. At the same time, it was shown that the robustness and generalizability of the models can be significantly increased by generating high-quality competitive examples and including them in the training sample [21].

Currently, some success has been achieved in creating a methodology for constructing adversarial examples in the field of image recognition [75] and audio processing [7]. However, the field of natural language word processing remains under-explored due to the specific nature of the input data. In addition to adversarial itself, such textual examples must have additional conditions:

- consistency of human interpretation - an adversarial example must be perceived by a person in the same way as the original one;
- semantic similarity - an adversarial example must carry the same semantic meaning for a person as the original one;
- linguistic correctness - an adversarial example must be perceived by a person as syntactically correct.

Existing methods of creating textual adversarial examples include misspelling words [39], removing individual words from text [40], deleting or inserting entire phrases [38], all of which generate artificial examples.

The statement of a method for generating adversarial examples for text classification problems can be formalized as follows [25]: having a set of sentences $X = x_1, x_2, ..., x_N$ and a corresponding set of labels $Y = y_1, y_2, ..., y_N$, a pre-trained model is given $F : X \rightarrow Y$ that assigns a label Y to a sentence X. In this case, for an arbitrary sentence, $x \in X$ adversarial example $x_{ADV}$ must satisfy the following conditions: $F(x_{ADV}) \neq F(x), Sim(x, x_{ADV}) \geq \epsilon$ where $Sim(\cdot)$ is the function of syntactic and semantic similarity, and $\epsilon$ is the minimum level of similarity between the original and adversarial examples.

If the black box condition is met, the adversarial example generation technique should not have information about the architecture, parameters, and structure of the training sample of the target machine learning model. She can only use the model on the given input example to obtain the classification result and confidence level.

J. Howard, J. Jijding, J. Tianijou, and P. Zolovits from MIT and Hong Kong University [25] proposed an algorithm for constructing adversarial textual examples TEXTFOOLER for studying black-box models, including BERT, consisting of the following steps:

- ranking the importance of words. The practice of using modern textual models shows that only some words serve as indicative factors in the operation of models, which is consistent with the studies of BERT T. Niven and H. Kao [56], which show the importance of the attention mechanism. Note that this step is trivial under the condition of the white box model and is reduced to calculating the gradient of the model output when changing individual words.
- transformation of words. In this step, words with the identified high importance for the model are replaced. The replacement must satisfy three conditions: have similar semantic meaning, be syntactically appropriate to the original context, and cause the model to mispredict. At this stage, several sub-stages can be distinguished:

- highlighting synonyms. A set of synonyms is selected based on the measure of the cosine distance between the original and all other words in the dictionary. Text embeddings at the word level can be used here [54].
- filtering by part of speech. To ensure syntactic consistency, it is necessary to leave only those candidate synonyms that coincide in grammatical role with the original word.
- checking semantic similarity. For each candidate, the similarity of the textual representation of the original sentence and the sentence in which the original word is replaced by it (the candidate for adversarial examples) is checked. For this, the USE model [9] is used, which represents a universal textual representation at the sentence level and a cosine similarity metric. It is necessary to achieve a similarity that exceeds a predetermined threshold $\epsilon$.
- validation of model prediction. For each candidate for adversarial examples, the prediction of the model is evaluated and, if examples are found that change the prediction of the model relative to the original proposal, they are added to the result set. If not, then the word with the next in descending order of importance, assessed in the first step, is selected and the process is repeated.

This algorithm was tested on five publicly available datasets for tact classification - AG's news [94], Fake news detection [98], MR [60], IMDB [99], and Yelp. Three modern text representation models were trained: WordCNN [30], WordLSTM [23] and BERT. The simulation results are shown in the table:

| | WordCNN | | | | | WordLSTM | | | | | BERT | | | | |
|---|---|---|---|---|---|---|---|---|---|---|---|---|---|---|---|
| | MR | IMDB | Yelp | AG | Fake | MR | IMDB | Yelp | AG | Fake | MR | IMDB | Yelp | AG | Fake |
| Original Accuracy | 78.0 | 89.2 | 93.8 | 91.5 | 96.7 | 80.7 | 89.8 | 96.0 | 91.3 | 94.0 | 86.0 | 90.9 | 97.0 | 94.2 | 97.8 |
| After-Attack Accuracy | 2.8 | 0.0 | 1.1 | 1.5 | 15.9 | 3.1 | 0.3 | 2.1 | 3.8 | 16.4 | 11.5 | 13.6 | 6.6 | 12.5 | 19.3 |
| % Perturbed Words | 14.3 | 3.5 | 8.3 | 15.2 | 11.0 | 14.9 | 5.1 | 10.6 | 18.6 | 10.1 | 16.7 | 6.1 | 13.9 | 22.0 | 11.7 |
| Semantic Similarity | 0.68 | 0.89 | 0.82 | 0.76 | 0.82 | 0.67 | 0.87 | 0.79 | 0.63 | 0.80 | 0.65 | 0.86 | 0.74 | 0.57 | 0.76 |
| Query Number | 123 | 524 | 487 | 228 | 3367 | 126 | 666 | 629 | 273 | 3343 | 166 | 1134 | 827 | 357 | 4403 |
| Average Text Length | 20 | 215 | 152 | 43 | 885 | 20 | 215 | 152 | 43 | 885 | 20 | 215 | 152 | 43 | 885 |

Source: [25]

When analyzing the results, it turned out that a dataset consisting of generated adversarial examples can reduce the efficiency of text classification from 80-97% to 0-20%. This indicates the success of the attack on the machine learning model.

Adding the generated adversarial dataset and additional training of the model on it significantly increases the efficiency of text classification models, which show 2-7 percentage points higher efficiency on the test adversarial dataset than without such additional training.

| | MR | | SNLI | |
|---|---|---|---|---|
| | Af. Acc. | Pert. | Af. Acc. | Pert. |
| Original | 11.5 | 16.7 | 4.0 | 18.5 |
| + Adv. Training | 18.7 | 21.0 | 8.3 | 20.1 |

Source: [25]

Undoubtedly, studies of the stability of machine learning models are an indispensable condition for the widespread introduction of such intelligent systems in the decision-making process, which is the relevance of this area of research. As mentioned earlier, the development and study of various types of automated attacks on machine learning models can give us a direction for further improving the development and testing of intelligent systems in general, not limited to classification models.

### Interlingual training of linguistic models

Generative pre-sentence encoders show their effectiveness in many natural text processing tasks [16,24,66]. Although the efforts of researchers are aimed at building a model for the general understanding of the text, most of the research is focused on monolingual representations trained on the corpus of English-language texts [11, 83]. Recent advances in teaching and evaluating interlanguage representations of sentences are aimed at overcoming the anglocentric bias and suggest that language models can learn truly universal representations in the common latent space of features, being trained on a corpus of texts in different languages. Such models belong to the group of interlanguage representations (XLU, [34]).

Along with monolingual studies, some works [68] show that the use of pre-trained language models can give an advantage in solving machine translation problems even in such classical language pairs as EN-DE, for which there is a large corpus of parallel texts, which facilitates the use of learning models with the teacher.

Equalization of distributions of sentence representations has been used for a long time in text nesting research. For example, T. Mikolov's work [52] uses short dictionaries to align representations for different languages. Further studies show that interlanguage representations can improve the quality of monolingual text embeddings [17], orthogonal transformations are sufficient to equalize the distribution of words between several languages [86], and this approach can be extended to an arbitrary number of languages [2]. The need to train such models with a teacher on pairs of parallel texts in different languages is reduced [71] and even eliminated [12].

A certain number of works in the field of searching for interlanguage representations are devoted to the problem of the so-called "zero-shot" problem - after training on a corpus of parallel texts, the model should adequately perform the tasks of text classification or translation into a pair of languages that it did not see clearly during its training. One of the most successful works in this area [26] presented a machine translation model based on a simple pair of encoder and decoder in LSTM. This model shows the best result among all machine translation models in rare language pairs, including new pairs that have not been studied in the learning process. The work of M. Artetze and others [3] shows that this model can serve as a basis for the creation of interlanguage textual representations. This approach uses training on over 200 million parallel sentences.

Recent work in the field of unsupervised learning shows that sentence representations can be aligned completely unsupervised [35]. Work [82] presents an LSTM model trained on sentences in different languages. The authors use common memory cell weights, but use different lookup tables to encode words in different languages. Their model shows good results on word translation problems.

The problem of building cross-language models (XLM) is discussed in detail in [34], where the authors investigate the process of training models on three learning tasks (conditional linguistic modeling, modeling of masked words, and modeling of text translation, CLM, MLM and TLM, respectively), two of which require training unsupervised on monolingual data, and the third is supervised learning on a set of parallel sentences.

The task of conditional modeling is to train the model to predict the probability of a word appearing in a sentence for a given set of previous words. Traditionally, this task is solved using recurrent neural networks, although the use of transformers [15] also shows its effectiveness.

Masked language modeling, also known as the Klose problem [76], is widely used to train text attachments and consists of teaching a model to predict a word (token) in a text sequence, replaced by a special token [MASK]. This technique is one of two BERT training tasks and is detailed in [16].

The text translation modeling task aims to use the parallel text corpus, when available, to improve the alignment of textual representations between languages. It is similar to the MLM problem with the difference that two parallel sentences in different languages are submitted as input. Words are masked in both sentences at random. To predict the masked word, the model can use its context in the original language, or, if there is not enough information, a second sentence, which is a translation of this sentence into another language. Thus, the model is stimulated to develop aligned textual representations between different languages. A graphical comparison of the last two tasks is shown in the figure:

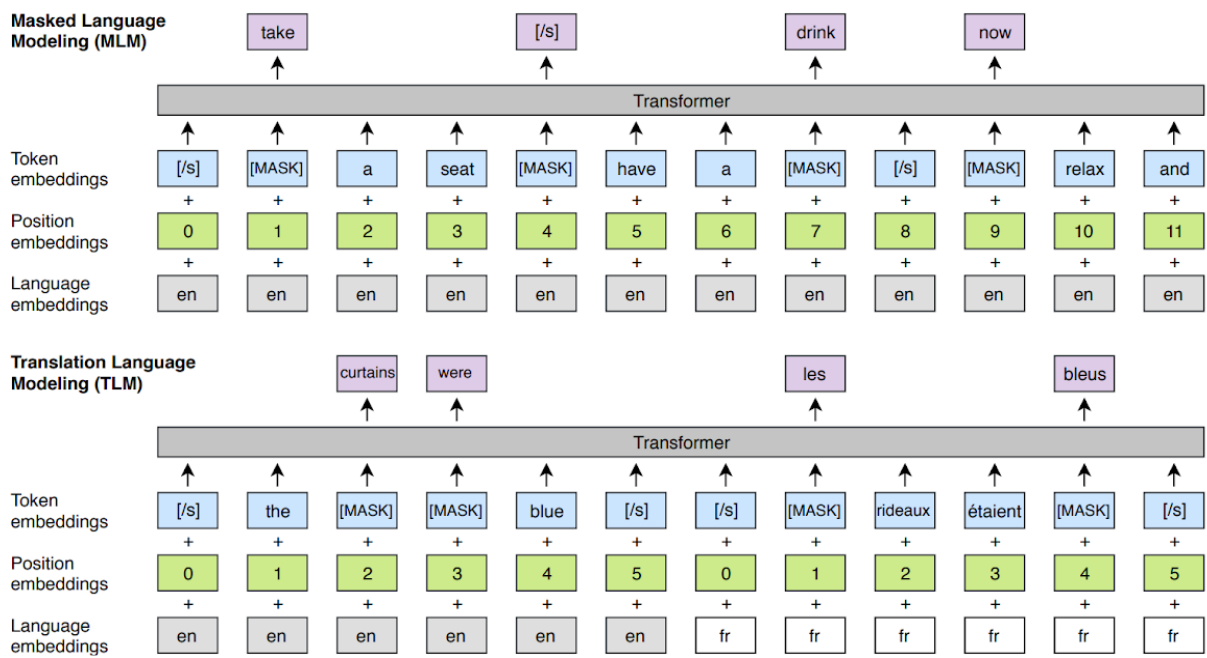

Source: [34]

To assess the quality of learning interlanguage models, the XLNI dataset [13] is used from the corpus of texts in 15 different languages, significantly differing in prevalence, from English to Nepali. Research [34] shows that the use of unsupervised multilingual learning can force the model to generate reliable cross-language textual representations. MLM training is especially effective. Without the use of parallel corpora, the model shows an improvement in the quality of machine translation by 1.3% on average relative to the best-known machine learning models. The TLM learning procedure can further improve machine translation accuracy by an average of 4.9% on average.

**Construction of domain-specific text models based on BERT The**

exponential growth in the number of scientific publications leads to the need to use automated word processing tools for large-scale knowledge extraction. The use of neural networks and deep learning is currently the most widespread, but such approaches require a large amount of pre-mapped text. For general language corpora, the availability of such data sets does not pose such a problem as for highly specialized scientific texts, where manual marking is associated with expensive expertise.

Modern ELMo, GPT, and BERT models show that using unsupervised learning can significantly improve model performance on a wide range of machine learning problems. Such models compute contextualized representations of words or text tokens that can be used as input to minimalistic domain-specific models. Such models are trained on large corpora of general-purpose texts, which makes them rather general and universal. However, when analyzing texts from a specific subject area, general models may show sub-optimal efficiency due to a lack of specific training.

In this regard, the emergence of some specialized models based on the architecture and teaching methodology of BERT, which has shown its effectiveness, but previously trained on specialized text corpora, is quite logical.

The authors of the BioBERT model (Seoul University), as the name implies, propose a specialized BERT-based text nesting model for the analysis of biomedical publications. The motivation for developing this model and the approach to teaching are completely similar.

The training took place according to the BERT method [16] on a corpus consisting of annotations of articles from the PubMed database and full-text versions of articles from the PMC database. The training took place for 23 days on 8 NVIDIA V100 GPUs based on weights exported from the original pretrained BERT model, that is, formally, this is additional training. Note that the total volume of the biomedical corpus was 18 billion words, while the volume for training the original model was only 3.3 billion. The learning process is schematically shown in the figure:

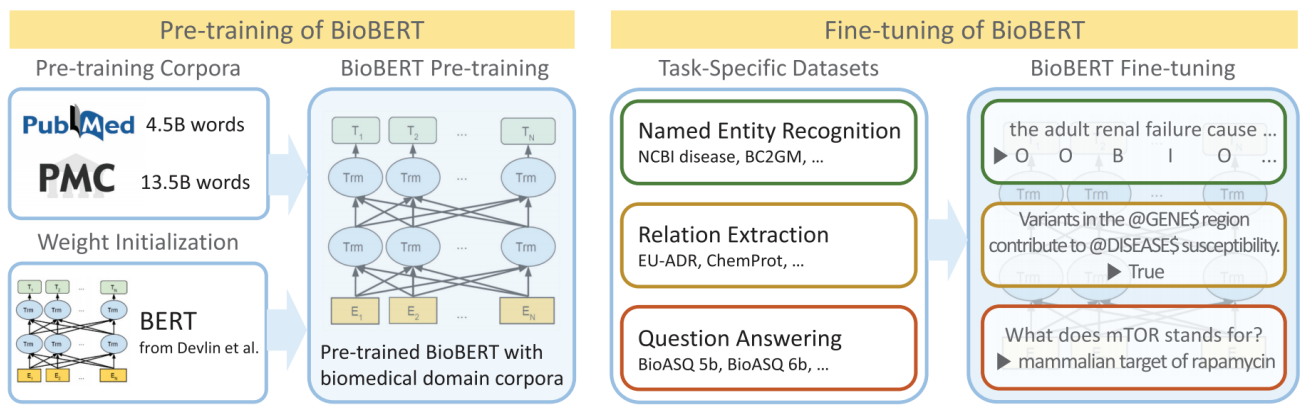

Source: [36]

Unlike SciBERT, the authors of this model did not use a specialized dictionary to tokenize text.

The BioBERT training process, like most similar models, consists of two parts: initial training and adjustment (additional training) for a specific word processing task. The authors of the model used three common tasks to assess the performance of the model: named entity recognition (NER [89]), relation extraction (REL), and answering a question (QA [67]).

The simulation showed the full advantage of the new model over conventional BERT in all specific tasks. The best results to date have been achieved in several tasks.

BioBERT is the first, but not the only problem-specific model based on BERT. I. Betalgi, K. Lo, and A. Cohan from the University of Seattle [6] present a SciBERT model trained on a corpus of scientific articles.

This model was trained on a random sample of more than 1 million scientific articles from the SemanticScholar database [1]. SciBERT also reconfigures the tokenization tool underlying BERT to account for the different vocabulary of the selected corpus. In the process of building the tokenizer, it turned out that the intersection of general-purpose dictionaries and the scientific corpus was 42%, which indicates a significant difference in the distribution of the token between these corpora. Along with an intuitive understanding, this fact gives reason to believe that the use of specific models can give increase productivity.

The model was trained following the original BERT training method, but on a different, more compact body of texts and using a specialized vocabulary for tokenization. The training took 7 days on an 8-core TPU (by comparison, training the full version of the original BERT model took 4 days on a 16-core TPU and is estimated to be 40-70 days on an 8-core).

As a result of numerical experiments on some tasks (extraction of entities, classification of texts, determination of dependencies, etc.) for the processing of scientific texts. Datasets from subject areas were used: biomedicine (ChemProt [31], EBM-NLP [57] and others), computer science (SciERC [47], ACL-ARC [28]), and interdisciplinary (SciCite, described by A. Cohan in [10 ]). On all datasets tested, SciBERT performed better than overall BERT and performed best to date for most.

| Field | Task | Dataset | SOTA | Bert-Base | | SciBert | |
|---|---|---|---|---|---|---|---|
| | | | | Frozen | Finetune | Frozen | Finetune |
| Bio | NER | BC5CDR (Li et al., 2016) | 88.85[7] | 85.08 | 86.72 | 88.73 | **90.01** |
| | | JNLPBA (Collier and Kim, 2004) | **78.58** | 74.05 | 76.09 | 75.77 | 77.28 |
| | | NCBI-disease (Dogan et al., 2014) | **89.36** | 84.06 | 86.88 | 86.39 | 88.57 |
| | PICO | EBM-NLP (Nye et al., 2018) | 66.30 | 61.44 | 71.53 | 68.30 | **72.28** |
| | DEP | GENIA (Kim et al., 2003) - LAS | **91.92** | 90.22 | 90.33 | 90.36 | 90.43 |
| | | GENIA (Kim et al., 2003) - UAS | **92.84** | 91.84 | 91.89 | 92.00 | 91.99 |
| | REL | ChemProt (Kringelum et al., 2016) | 76.68 | 68.21 | 79.14 | 75.03 | **83.64** |
| CS | NER | SciERC (Luan et al., 2018) | 64.20 | 63.58 | 65.24 | 65.77 | **67.57** |
| | REL | SciERC (Luan et al., 2018) | n/a | 72.74 | 78.71 | 75.25 | **79.97** |
| | CLS | ACL-ARC (Jurgens et al., 2018) | 67.9 | 62.04 | 63.91 | 60.74 | **70.98** |
| Multi | CLS | Paper Field | n/a | 63.64 | 65.37 | 64.38 | **65.71** |
| | | SciCite (Cohan et al., 2019) | 84.0 | 84.31 | 84.85 | **85.42** | 85.49 |
| Average | | | | 73.58 | 77.16 | 76.01 | 79.27 |

Table 1: Test performances of all Bert variants on all tasks and datasets. **Bold** indicates the SOTA result (multiple results bolded if difference within 95% bootstrap confidence interval). Keeping with past work, we report macro F1 scores for NER (span-level), macro F1 scores for REL and CLS (sentence-level), and macro F1 for PICO (token-level), and micro F1 for ChemProt specifically. For DEP, we report labeled (LAS) and unlabeled (UAS) attachment scores (excluding punctuation) for the same model with hyperparameters tuned for LAS. All results are the average of multiple runs with different random seeds.

Source: [6]

It is noteworthy that this model shows comparable and sometimes superior results compared to BioBERT [36], despite being trained on a significantly smaller body of biomedical texts.

The authors of the SciBERT model separately note that in all cases, additional training of text attachments on a specific task gives a greater effect than building special architectures based on fixed attachments. Also, the use of a vocabulary built for the subject area for text tokenization gives a positive effect.

**Investigation of the robustness of BERT learning**

Despite the rapid and successful spread of BERT in almost all areas of word processing in natural languages, there are no guarantees that the presented pre-trained model is optimal, both in terms of architecture and in terms of training methodology and text corpora used. I. Liu and colleagues from the Facebook AI group [43] analyzed the robustness of the BERT model and came to some interesting conclusions.

Following, to a large extent, the original procedure for training the model, they concentrated on obtaining the largest possible open corpus of texts for the formation of the training sample. Initially, BERT is trained on a collection of BookCorpus [96] and Wikipedia, which is about 16 GB of uncompressed text. In this work, we used four additional public corpora of English-language texts - CC-News [100], OpenWebText [58], and Stories [78], totaling about 160 GB, that is, an order of magnitude more than the original.

After training, the resulting model was tested against three common benchmarks measuring natural text understanding - GLUE [83], SQuAD [67], and RACE [33]. On all these tasks, the authors received an improvement in the initial indicators, which allows us to conclude that the BERT model is undertrained.

In the course of training, some modifications of the original method were analyzed and factor analysis of their influence on the overall performance of the model on the three above-mentioned tasks was made.

For example, in the process of preparing a training sample for the masked language model BERT problem, the input text tokens are masked once at the preprocessing stage, that is, static text masking is implemented. Thus, the model sees the same sentence with the same masks at all learning epochs. In contrast, one can duplicate the original sentence multiple times and apply randomized masking for each learning epoch. This is called dynamic text masking. It turns out that such dynamic masking can slightly increase the robustness and efficiency of the resulting model.

During the BERT training process, the importance of the second training task, that is, the next sentence prediction (NSP), was revealed. The authors of [16] note that removing this loss function from training significantly reduces performance. However, more recent studies [27,34,87] question the need for NSP inclusion. Several training options for this task were tested to address this issue. First, the original method of two segments, each of which could contain several sentences of the original text, but both together do not exceed 512 tokens in length. Second, the use of a pair of complete sentences selected either sequentially from one text, or different texts in the corpus. Thirdly, training without NSP, when whole related sentences from one or several documents are submitted for input. And fourthly, the same thing, but with the prohibition of sentences from several documents. The learning outcomes are shown in Figure

| Model | SQuAD 1.1/2.0 | MNLI-m | SST-2 | RACE |
|---|---|---|---|---|
| *Our reimplementation (with NSP loss):* | | | | |
| SEGMENT-PAIR | 90.4/78.7 | 84.0 | 92.9 | 64.2 |
| SENTENCE-PAIR | 88.7/76.2 | 82.9 | 92.1 | 63.0 |
| *Our reimplementation (without NSP loss):* | | | | |
| FULL-SENTENCES | 90.4/79.1 | 84.7 | 92.5 | 64.8 |
| DOC-SENTENCES | 90.6/79.7 | 84.7 | 92.7 | 65.6 |
| BERT$_{\text{BASE}}$ | 88.5/76.3 | 84.3 | 92.8 | 64.3 |
| XLNet$_{\text{BASE}}$ ($K = 7$) | –/81.3 | 85.8 | 92.7 | 66.1 |
| XLNet$_{\text{BASE}}$ ($K = 6$) | –/81.0 | 85.6 | 93.4 | 66.7 |

Source: [43]

When testing the training, it was concluded that removing the NSP function improves performance on subsequent tasks, which directly contradicts the original publication. Also, using sentences instead of segments degrades performance, presumably by removing the ability to make long-term generalizations.

Initially, BERT was trained in 1 million steps in bursts of 256 sequences. This is computationally equivalent to training in 30 thousand steps but large packets of 8 thousand sequences. There are studies by M. Ott et al. [59]showing that language models can benefit from an increase in the size of the training packet, provided the learning rate is increased accordingly. Research [43] shows that this effect also takes place for BERT, but to a limited extent; the highest performance is achieved with a packet size of about 2 thousand sequences.

A scrupulous study of all the intricacies of learning a linguistic model, given in [43], led to the creation of a variation of the model called RoBERTa (robustly optimized approach to BERT), which combines the best practices from those analyzed. As a result, it was found that the performance of the original model can be improved by an average of 3-4 percentage points, reaching the best values to date in all tested problems.

## Conclusion

Immediately after its appearance, the BERT model received an intense reaction from the scientific community and is now used in almost all word processing problems. Almost immediately, the proposals for improving the model considered in this work appeared, which led to an improvement in the results of its application in all subsequent problems. With all that said, we can confidently assert that BERT represented a quantum leap in the field of intelligent natural language processing and consolidated the superiority of using pre-trained text representation models on huge data sets as a universal basis for building intelligent algorithms for solving specific problems.

BERT has also shown the advantage of bidirectional contextual models of text comprehension based on the architecture of transformers with an attention mechanism.

Undoubtedly, we will see many more new scientific results based on the application and adaptation of the BERT model to various problems of word processing in natural languages. Further improvement of the neural network architecture, coupled with fine-tuning the training procedure and parameters, will inevitably lead to significant improvements in many computer NLP algorithms, from text classification and annotation to machine translation and question-answer systems.